\documentclass{article}
\usepackage{spconf,amsmath,graphicx}
\usepackage{booktabs}
\usepackage{enumitem}
\usepackage[T1]{fontenc}
\usepackage{times}
\title{Lesion Overlap Ranks PET Segmentation Models but Does Not Certify Patient-Level Tumor Burden}

\name{\textbf{Tushar Kataria}$^{1,2}$
\quad
\textbf{Gerald Sabin}$^{3}$
\quad
\textbf{Ponnuswamy Sadayappan}$^{2}$
\quad
\textbf{Shireen Y. Elhabian}$^{1,2}$}
\address{$^{1}$ Scientific Computing and Imaging Institute, University of Utah, Salt Lake City, UT, USA \\
$^{2}$ Kahlert School of Computing, University of Utah, Salt Lake City, UT, USA  \\
$^{3}$ RNET Technologies, Dayton, OH, USA\\
{\tt\small tushar.kataria@utah.edu,gsabin@rnet-tech.com, saday@cs.utah.edu, shireen@sci.utah.edu}
}

\begin{document}
\maketitle

\begin{abstract}
Metabolic tumor volume (MTV) summarizes whole-body PET disease burden for prognostic stratification and treatment assessment. Automated MTV is useful only if its estimate is reliable for the individual patient, yet segmentation models are usually evaluated with overlap metrics such as Dice rather than burden error. We examined whether overlap metrics identify unreliable MTV estimates using 1,014 autoPET FDG PET/CT studies. We evaluated LesionTracer predictions held out from fine-tuning, compared five selected released models on 489 tumor-positive patients, and examined the findings in 597 PSMA studies. Across the five FDG models, ranking by higher mean Dice gave the same order as ranking by lower mean relative MTV error ($\rho=-1.00$, exact $p=0.017$). At the patient level, however, out-of-fold Dice tracked relative MTV error strongly ($\rho=-0.75$), yet 82 of 489 patients (16.8\%) fell on opposite sides of the Dice 0.80 and MTV-error 25\% thresholds. Within the training-exposed released models, correlations ranged from $-0.47$ to $-0.74$ over narrower, high-Dice ranges. Out of fold, opposing false-positive and false-negative volumes cancelled about one third of spatial error in net MTV. In 236 of 489 patients, each of the five models missed at least one lesion, though not necessarily the same lesion. Thus, agreement in average model rankings did not establish the reliability of individual MTV estimates or reveal shared lesion misses. PET segmentation benchmarks should report patient-level MTV error and missed-lesion counts alongside Dice and other overlap metrics.
\end{abstract}


\section{Introduction}
\label{sec:intro}

Whole-body PET enables disease burden to be measured across multiple lesions. Metabolic tumor volume (MTV) and total lesion glycolysis (TLG) summarize this burden and have been studied for prognostic stratification and treatment assessment~\cite{you2026metabolic,mikhaeel2022proposed,chung2014fdg,mcdonald2017assessment}. Automated lesion segmentation could make these measurements available at scale~\cite{gatidis2024results,dexl2026autopet3}, but missed tumor, inaccurate boundaries, or physiological false-positive uptake can distort an individual patient's estimate. This distinction matters because prognostic analyses use the numerical MTV estimate to stratify patients by risk~\cite{mikhaeel2022proposed,chung2014fdg}.

Segmentation benchmarks commonly compare models using overlap measures such as Dice and intersection-over-union (IoU)~\cite{maier2024metrics,reinke2024understanding,kataria2023pretrain,li2024abdomenatlas,khan2024viability}. The autoPET challenges also measure the volume of entirely missed reference lesions and unmatched predicted components~\cite{gatidis2024results,dexl2026autopet3}. None of these challenge endpoints directly measures patient-level MTV error. Dice reflects the combined false-positive (FP) and false-negative (FN) voxel volume, whereas MTV error reflects their difference: FP and FN can cancel in the reported burden despite substantial segmentation error.

The limitations of overlap for quantitative PET are well recognized. The RELAINCE guidelines recommend evaluating AI methods against their intended clinical task~\cite{jha2022nuclear}. For primary lung tumors, Liu et al.~\cite{liu2024need} showed that Dice and MTV or TLG accuracy can favor different methods. In lymphoma, Yousefirizi et al.~\cite{yousefirizi2022novel} investigated how segmentation loss affects TMTV computation, while Ahamed et al.~\cite{ahamed2026comprehensive} measured patient-level errors in clinical lesion measures on multiple cohorts, including an external autoPET lymphoma cohort, and found that small, faint lesions remain difficult to detect. What remains unclear on a common whole-body benchmark is whether overlap identifies unreliable MTV estimates for individual patients and whether released models miss the same lesions. We address these questions through a patient-level audit of autoPET~\cite{gatidis2022whole}: we reconstruct LesionTracer's validation folds~\cite{rokuss2024fdg} to obtain predictions held out from fine-tuning, compare five released models selected for this study on the same FDG patients, and examine the main findings in PSMA PET/CT. Our contributions are:
\begin{itemize}[leftmargin=*, itemsep=0pt, parsep=0pt, topsep=0pt]
    \item We compare overlap with patient-level MTV error across out-of-fold and released-model analyses, including the effect of FP/FN cancellation.
    \item We identify lesions missed across models, characterize their size and uptake, and examine whether the findings persist in PSMA PET/CT.
\end{itemize}
\section{Study Design and Methods}
\label{sec:methods}

\textbf{Study design.} We evaluated the gap between segmentation overlap and MTV accuracy in three settings. LesionTracer\cite{rokuss2024fdg} predictions held out from its autoPET~III fine-tuning folds assess patient-level error, subject to the pretraining exposure described below. Five released FDG models allow us to compare model rankings and identify shared failures. A PSMA cohort tests whether the main patterns persist across tracers.

\textbf{Data.} We used the public FDG-UKT NIfTI release of autoPET~\cite{gatidis2022whole}: 1,014 whole-body [$^{18}$F]FDG PET/CT studies from 900 patients, comprising 501 tumor-positive and 513 tumor-negative studies. The release provides PET images in SUV, CT resampled to PET resolution, and binary lesion masks.

\textbf{Out-of-fold predictions.} LesionTracer, the autoPET~III winner~\cite{rokuss2024fdg}, is released as five fold checkpoints without a split file. We reconstructed fold membership from the per-case entries in each fold's validation summary. The validation sets were disjoint, covered the cohort, and were patient-disjoint, so each study was predicted by a checkpoint that had not seen that patient during autoPET~III fine-tuning. This is not a fully unexposed evaluation: the supervised foundation pretraining used to initialize LesionTracer included the same 1,014-study FDG cohort with lesion labels~\cite{ulrich2025large}. Thus, ``out of fold'' refers only to fine-tuning; these results do not estimate performance on patients absent from both pretraining and fine-tuning.

\textbf{Multi-model panel.} Fold membership could not be recovered for the other released models. We therefore evaluated five models as distributed: LesionTracer (autoPET~III rank~1)~\cite{rokuss2024fdg}, IKIM (rank~2)~\cite{kalisch2024autopet}, HussainAlasmawi (rank~3)~\cite{alasmawi2026advanced}, CW nnU-Net~\cite{wang2024dual}, and the top-ranked autoPET II entry, BAMF~\cite{murugesan2023improving}. All five had trained on the FDG cohort, so this comparison is reported separately from the out-of-fold analysis and does not measure performance on unseen patients. To examine whether the findings extend to another tracer, we used the 597-study PSMA release (378 patients) and reconstructed LesionTracer's validation folds in the same way.

\textbf{Endpoints.} For each tumor-positive study, we computed MTV from the lesion-mask volume and defined relative MTV error as $\varepsilon_{\mathrm{MTV}}=|\mathrm{MTV}_{\mathrm{pred}}-\mathrm{MTV}_{\mathrm{ref}}|/\mathrm{MTV}_{\mathrm{ref}}$. We defined relative total lesion glycolysis (TLG) error analogously, using SUV-weighted lesion volume. We also measured absolute MTV error in milliliters and bounded MTV error, $|\mathrm{MTV}_{\mathrm{pred}}-\mathrm{MTV}_{\mathrm{ref}}|/(\mathrm{MTV}_{\mathrm{pred}}+\mathrm{MTV}_{\mathrm{ref}})$. A study was classified as an MTV failure when $\varepsilon_{\mathrm{MTV}}>25\%$.

To examine how segmentation errors affect MTV, we measured false-positive (FP) and false-negative (FN) voxel volumes. Their difference determines the signed MTV error, $\mathrm{MTV}_{\mathrm{pred}}-\mathrm{MTV}_{\mathrm{ref}}=\mathrm{FP}-\mathrm{FN}$. We quantified cancellation as the fraction of total FP and FN volume absent from the absolute MTV error: $[\mathrm{FP}+\mathrm{FN}-|\mathrm{FP}-\mathrm{FN}|]/(\mathrm{FP}+\mathrm{FN})$. Cancellation includes FP and FN voxels from a shifted boundary within the same lesion, as well as errors at different locations. We additionally measured the summed absolute volume discrepancy on detected lesions, the volume of entirely missed lesions, and the volume of predicted components that overlapped no reference lesion. Reference lesions were defined as 26-connected components; a lesion was missed if no predicted voxel overlapped it. We characterized each lesion by its physical 3D Feret diameter and mean SUV. Predicted components overlapping multiple lesions were partitioned by nearest distance.

\textbf{Statistics.} We computed endpoints for each study and averaged them across studies from the same patient, so the patient was the unit of analysis (489 tumor-positive FDG patients; 368 PSMA patients). We used Spearman correlations, partial Spearman correlations (rank residuals after regressing on reference MTV, lesion count, and small-lesion fraction), and patient-level bootstrap confidence intervals (5,000 resamples). Paired comparisons used Wilcoxon tests with Holm correction. We assessed the five-model ranking correlation with an exact permutation test. Only nine patients had more than one tumor-positive study; using each patient's first study did not significantly change any endpoint (all paired $p>0.05$).

\section{Results and Discussion}
\label{sec:results}

\textbf{Overlap ranks models, but does not validate individual estimates.} Out of fold, mean (median) patient Dice was 0.773 (0.834), and mean (median) $\varepsilon_{\mathrm{MTV}}$ was 37.5\% (17.6\%); 39.1\% of patients exceeded 25\% MTV error (Table~\ref{tab:models}). The five released models selected for this study achieved higher Dice scores, although their evaluation was training-exposed: each five-fold ensemble combined one checkpoint that held out the study with four that trained on it, while CW used a single model trained on all cases. Nevertheless, at most 16.6\% of patients reached Dice 0.95 under any model.

Across the five released models selected for this study, mean Dice ranked mean $\varepsilon_{\mathrm{MTV}}$ in exactly reverse order ($\rho=-1.00$, exact $p=0.017$; Table~\ref{tab:models}). The models spanned a mean Dice range of 0.79--0.90, over which overlap distinguished their average burden accuracy. This agreement is specific to the evaluated models: Liu et al.~\cite{liu2024need} found that Dice and MTV or TLG accuracy can favor different methods. Moreover, agreement in model rankings does not show whether Dice identifies an unreliable MTV estimate for an individual patient.

\begin{table}[!htb]
\centering
\caption{Patient-level results for 489 tumor-positive FDG patients. OOF: LesionTracer predictions held out from fine-tuning; other rows are training-exposed released models. Ranks refer to autoPET~III (III) or autoPET~II (II); the evaluated Hussain model combines two released configurations. $\varepsilon_{\mathrm{TLG}}$: mean relative TLG error. Patients with miss (\%): patients with at least one entirely missed lesion. $\rho$: within-model Spearman correlation between patient Dice and $\varepsilon_{\mathrm{MTV}}$.}
\label{tab:models}
\setlength{\tabcolsep}{1pt}
\resizebox{\linewidth}{!}{%
\begin{tabular}{lccccc}
\toprule
Model (rank) & Dice & $\varepsilon_{\mathrm{MTV}}$ mean (med) & $\varepsilon_{\mathrm{TLG}}$ & \% Missed & $\rho$ \\
\midrule
LesionTracer OOF (III-1) & 0.773 & 37.5\% (17.6\%) & 35.7\% & 67.1\% & $-0.75$ \\
\midrule
CW                    & 0.896 & 6.6\% (4.7\%)   & 4.0\%  & 59.5\% & $-0.47$ \\
LesionTracer (III-1)  & 0.895 & 7.3\% (4.8\%)   & 6.0\%  & 57.5\% & $-0.49$ \\
IKIM (III-2)          & 0.877 & 10.3\% (7.0\%)  & 7.8\%  & 53.0\% & $-0.56$ \\
Hussain (III-3)       & 0.848 & 14.8\% (9.3\%)  & 10.4\% & 65.2\% & $-0.68$ \\
BAMF (II-1)           & 0.793 & 35.6\% (17.2\%) & 49.9\% & 60.5\% & $-0.74$ \\
\bottomrule
\end{tabular}}
\end{table}

\begin{table}[!htb]
\centering
\caption{Error composition for the five released models selected for this study (489 patients). Values are patient means except missed-lesion counts. FP and FN are false-positive and false-negative voxel volumes; $P$ and $R$ are predicted and reference volumes. For each patient, $e=(\mathrm{FP}+\mathrm{FN})/(P+R)=1-\mathrm{Dice}$. Canc.: fraction of FP\,+\,FN volume cancelled in net MTV. Missed lesions (n): total number of entirely missed reference lesions.}
\label{tab:errors}
\setlength{\tabcolsep}{1pt}
\resizebox{\linewidth}{!}{%
\begin{tabular}{lccccccc}
\toprule
Model & Dice & FP (mL) & FN (mL) & FP\,+\,FN (mL) & $e$ & Canc. & Missed Lesion(n) \\
\midrule
CW           & \textbf{0.896} & 18.7          & 25.9          & 44.6          & \textbf{0.104} & 0.67 & 2,657 \\
LesionTracer & 0.895          & 16.3          & 25.8          & \textbf{42.1} & 0.105          & 0.67          & 2,271 \\
IKIM         & 0.877          & 21.2          & \textbf{24.9} & 46.1          & 0.123          & 0.60          & \textbf{1,948} \\
Hussain      & 0.848          & \textbf{15.8} & 32.6          & 48.4          & 0.152          & 0.51          & 2,592 \\
BAMF         & 0.793          & 23.8          & 45.8          & 69.6          & 0.206          & 0.42          & 2,923 \\
\bottomrule
\end{tabular}}
\end{table}

This ranking agreement is empirical, not a property of Dice. Dice $=1-(\mathrm{FP}+\mathrm{FN})/(P+R)$ reflects total spatial error relative to predicted and reference volume, whereas signed MTV error equals $\mathrm{FP}-\mathrm{FN}$. Accordingly, the highest-Dice model did not have the least FP volume (Hussain), the least FN volume or fewest missed lesions (IKIM), or the least combined FP and FN volume (LesionTracer; Table~\ref{tab:errors}). The two highest-Dice models also had the greatest FP/FN cancellation (0.67 versus 0.42 for BAMF). Dice can rank models by mean MTV error in this cohort without directly measuring the errors that determine an individual patient's MTV.

\textbf{Patient-level Dice cannot certify MTV.} Out of fold, patient Dice was strongly associated with relative MTV error ($\rho=-0.75$, 95\% CI $-0.80$ to $-0.70$), including after adjustment for reference MTV, lesion count, and small-lesion fraction (partial $\rho=-0.71$). Across the training-exposed released models, the association was weaker ($\rho=-0.47$ to $-0.74$; Table~\ref{tab:models}), partly in a more restricted, high-Dice range. Dice was nearly unrelated to absolute MTV error in milliliters out of fold ($\rho=-0.08$). Relative and absolute error also identified different patients: among the 25 patients with the largest error on each scale, only one appeared in both groups. The error scale therefore matters when using overlap to assess an individual burden estimate.

\begin{figure}[!htb]
  \centering
  \includegraphics[width=\linewidth]{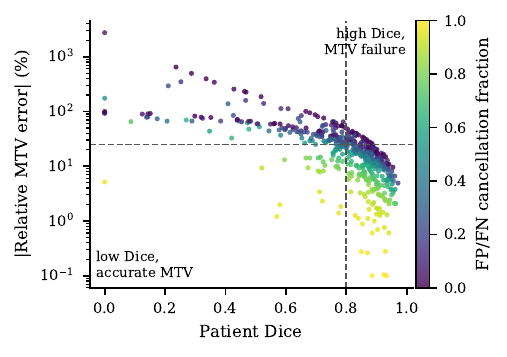}
  \vspace{-2em}
  \caption{Out-of-fold LesionTracer: patient Dice versus $\varepsilon_{\mathrm{MTV}}$ (489 patients), coloured by FP/FN cancellation fraction. Dashed lines mark Dice 0.80 and 25\% MTV error.}
  \label{fig:oof}
\end{figure}

A Dice threshold makes the mismatch visible (Fig.~\ref{fig:oof}). At Dice 0.80 and 25\% MTV error, 82 of 489 patients (41 in each direction) fell on opposite sides of the two thresholds. Patients with high Dice but inaccurate MTV tended to have more FP than FN volume. Those with low Dice but accurate MTV often had FP and FN volumes that cancelled in the total burden. Across all patients, 24.3 of the mean 71.9\,mL FP\,+\,FN volume cancelled in net MTV. Thus, an accurate total MTV could reflect opposing segmentation errors rather than accurate lesion delineation.

\begin{figure}[!htb]
  \centering
  \includegraphics[width=\linewidth]{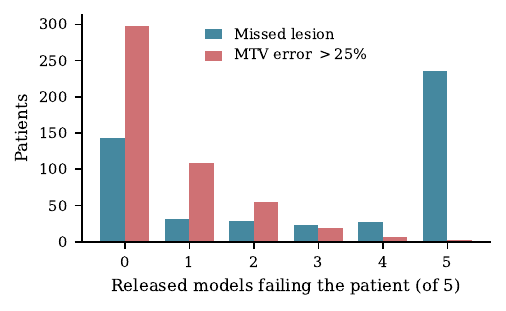}
    \vspace{-2em}
  \caption{Number of models that missed at least one lesion in each patient, compared with the number producing MTV error $>$25\% (n = 489).}
  \label{fig:panel}
\end{figure}

\textbf{Released models share a small-lesion blind spot.} In 236 of 489 patients, each of the five evaluated models missed at least one lesion; the missed lesions need not be the same across models (Fig.~\ref{fig:panel}). Only three patients exceeded 25\% MTV error under every model. At the lesion level, 1,620 of 8,620 reference lesions were missed by all five models. Of these shared misses, 90.5\% were smaller than 15\,mm and had SUV\textsubscript{mean} below 5.

In LesionTracer's out-of-fold predictions, the median errors from matched-lesion volume differences, missed lesions, and isolated false-positive components were 15.7\%, 0.3\%, and 0.4\% of reference MTV, respectively. Matched-lesion error accounted for a median 88\% of each patient's summed component error. Thus, frequent lesion misses contributed relatively little volume compared with errors on detected lesions.

\begin{figure}[!htb]
  \centering
  \includegraphics[width=\linewidth]{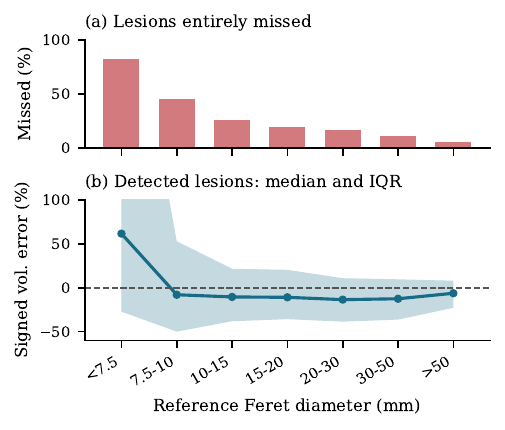}
   \vspace{-2em}
  \caption{Out-of-fold LesionTracer by reference Feret diameter. (a) Fraction of reference lesions entirely missed. (b) Signed relative volume error of detected lesions; the band shows the interquartile range, clipped at 100\%.}
  \label{fig:size}
\end{figure}

\textbf{Small lesions expose the limits of total burden.} Out of fold, lesions below 15\,mm were missed more often than lesions above 30\,mm (54.9\% versus 8.8\%). Among patients with small lesions, error in small-lesion burden exceeded error in total MTV by a median 31.5 percentage points (95\% CI 22.8--37.2; one-sided Wilcoxon $p=2.8\times10^{-31}$). All five evaluated released models showed the same pattern. Large lesions can dominate total MTV, leaving poorly quantified small-lesion disease hidden in an apparently accurate total. Lesions below 7.5\,mm were missed 82.4\% of the time; the few detected lesions were over-segmented (median +62\%). This pattern is consistent with the difficulty of delineating lesions near PET's spatial resolution, including partial-volume effects~\cite{soret2007partial}. From 7.5\,mm upward, detected lesions showed a comparatively stable underestimation of 6--14\% (Fig.~\ref{fig:size}).

\textbf{The patient-level gap persists in PSMA PET/CT.} On 597 PSMA studies (368 tumor-positive patients), out-of-fold LesionTracer predictions had lower overlap than on FDG (median Dice 0.67) and larger relative MTV error (median 33.8\%; 63.3\% of patients above 25\%). Dice tracked relative MTV error ($\rho=-0.81$, 95\% CI $-0.84$ to $-0.77$), including after adjustment for reference MTV, lesion count, and small-lesion fraction (partial $\rho=-0.78$). Nevertheless, 74 patients (20.1\%) fell on opposite sides of the Dice 0.80 and MTV-error 25\% thresholds. Most had low Dice but accurate net MTV (62 patients), while 12 had high Dice but inaccurate MTV. MTV was underestimated in 73.1\% of patients, with median FN volume exceeding FP volume (10.6 versus 3.0\,mL). Entirely missed lesions were common (68.8\% of patients), particularly among lesions below 15\,mm, and small-lesion error exceeded total MTV error in 194 of 296 affected patients (median difference +9.6 percentage points, $p=1.4\times10^{-9}$). Unlike FDG, discordance in PSMA was dominated by low Dice despite accurate net MTV, consistent with opposing segmentation errors cancelling in the reported burden.

\textbf{Limitations.} Only LesionTracer's validation folds could be reconstructed. Its supervised foundation pretraining included this same 1,014-study FDG cohort with lesion labels, so ``out of fold'' refers only to its released fine-tuning folds and may overestimate performance on previously unseen patients. The five-model FDG comparison used training-exposed models with different training data and architectures, so it compares released models rather than estimating performance on new patients. The out-of-fold PSMA analysis used one model, and each tracer cohort came from a single center; tracer and center effects cannot be separated. Some small shared misses may be reference-annotation fragments, although shared misses persisted for lesions $\geq$1\,mL. Shared misses do not identify their cause. Finally, the Dice 0.80 and MTV-error 25\% thresholds illustrate discordance and are not proposed clinical decision limits.

\section{Conclusion and Future Work}
\label{sec:conclusion}

Mean Dice ranked the five FDG models by mean MTV error, but did not establish the accuracy of an individual patient's estimate. Out of fold, Dice tracked relative MTV error, yet 82 of 489 patients fell on opposite sides of the Dice and MTV-error thresholds. Dice was weakly related to absolute error in milliliters, while FP/FN cancellation could make net MTV appear accurate despite spatial error. All five models missed many of the same small, low-uptake lesions, whose contribution to total MTV was often limited. Patient-level discordance also persisted in PSMA PET/CT. Future work should test whether these errors alter treatment-response or prognostic assessments and measure their frequency in patients without training exposure. We recommend reporting patient-level relative MTV error and missed-lesion counts alongside Dice in PET segmentation evaluations, with small, low-uptake lesions prioritized for expert review.

\section{Compliance with Ethical Standards}  
This retrospective study analyzed publicly released, anonymized autoPET FDG-UKT and PSMA PET/CT data. The authors had no access to patient identifiers and did not contact participants. Publication of the source datasets was approved by the respective institutional ethics committees.

\section{Acknowledgments} This work was supported by the National Science Foundation under Grant NSF2217154.  The content is solely the responsibility of the authors and does not necessarily represent the official views of the National Science Foundation. AI tools(ChatGPT and Claude Code) were used to assist in revising the text in the paper, writing code for inference scripts, confirming whether number reported match the experimental results and creating latex formatted tables. The authors reviewed the final text and independently verified the reported analyses and findings. The authors have no relevant financial or non-financial interests to disclose.

{
\scriptsize
\bibliographystyle{IEEEbib}
\bibliography{refs}

@article{you2026metabolic,
  title={Metabolic tumor volume and total lesion glycolysis for predicting prognosis in diffuse large B-cell lymphoma: a retrospective PET/CT study},
  author={You, Yang and Zhang, Weifeng and Zhen, Zhifei and Xu, Junling and Xuan, Ang},
  journal={Frontiers in Oncology},
  volume={16},
  pages={1862170},
  year={2026},
  publisher={Frontiers Media SA}
}

@article{mikhaeel2022proposed,
  title={Proposed new dynamic prognostic index for diffuse large B-cell lymphoma: international metabolic prognostic index},
  author={Mikhaeel, N George and Heymans, Martijn W and Eertink, Jakoba J and De Vet, Henrica CW and Boellaard, Ronald and D{\"u}hrsen, Ulrich and Ceriani, Luca and Schmitz, Christine and Wiegers, Sanne E and H{\"u}ttmann, Andreas and others},
  journal={Journal of Clinical Oncology},
  volume={40},
  number={21},
  pages={2352--2360},
  year={2022},
  publisher={Wolters Kluwer Health}
}

@article{ulrich2025large,
  title={Large scale supervised pretraining for traumatic brain injury segmentation},
  author={Ulrich, Constantin and Wald, Tassilo and Isensee, Fabian and Maier-Hein, Klaus H},
  journal={arXiv preprint arXiv:2504.06741},
  year={2025}
}

@article{soret2007partial,
  title={Partial-volume effect in PET tumor imaging},
  author={Soret, Marine and Bacharach, Stephen L and Buvat, Irene},
  journal={Journal of nuclear medicine},
  volume={48},
  number={6},
  pages={932--945},
  year={2007},
  publisher={Society of Nuclear Medicine}
}

@article{liu2024need,
  title={Need for objective task-based evaluation of image segmentation algorithms for quantitative PET: a study with ACRIN 6668/RTOG 0235 multicenter clinical trial data},
  author={Liu, Ziping and Mhlanga, Joyce C and Xia, Huitian and Siegel, Barry A and Jha, Abhinav K},
  journal={Journal of Nuclear Medicine},
  volume={65},
  number={3},
  pages={485--492},
  year={2024},
  publisher={Society of Nuclear Medicine}
}

@article{jha2022nuclear,
  title={Nuclear medicine and artificial intelligence: best practices for evaluation (the RELAINCE guidelines)},
  author={Jha, Abhinav K and Bradshaw, Tyler J and Buvat, Ir{\`e}ne and Hatt, Mathieu and Kc, Prabhat and Liu, Chi and Obuchowski, Nancy F and Saboury, Babak and Slomka, Piotr J and Sunderland, John J and others},
  journal={Journal of Nuclear Medicine},
  volume={63},
  number={9},
  pages={1288--1299},
  year={2022},
  publisher={Society of Nuclear Medicine}
}

@article{khan2024viability,
  title={On the Viability of Semi-Supervised Segmentation Methods for Statistical Shape Modeling},
  author={Khan, Asma and Kataria, Tushar and Ukey, Janmesh and Elhabian, Shireen Y},
  journal={arXiv preprint arXiv:2407.15260},
  year={2024}
}

@article{li2024abdomenatlas,
  title={AbdomenAtlas: A large-scale, detailed-annotated, \& multi-center dataset for efficient transfer learning and open algorithmic benchmarking},
  author={Li, Wenxuan and Qu, Chongyu and Chen, Xiaoxi and Bassi, Pedro RAS and Shi, Yijia and Lai, Yuxiang and Yu, Qian and Xue, Huimin and Chen, Yixiong and Lin, Xiaorui and Tang, Yutong and Cao, Yining and Han, Haoqi and Zhang, Zheyuan and Liu, Jiawei and Zhang, Tiezheng and Ma, Yujiu and Wang, Jincheng and Zhang, Guang and Yuille, Alan and Zhou, Zongwei},
  journal={Medical Image Analysis},
  volume={97},
  pages={103285},
  year={2024},
  publisher={Elsevier},
  url={https://github.com/MrGiovanni/AbdomenAtlas}
}

@inproceedings{kataria2023pretrain,
  title={To pretrain or not to pretrain? A case study of domain-specific pretraining for semantic segmentation in histopathology},
  author={Kataria, Tushar and Knudsen, Beatrice and Elhabian, Shireen},
  booktitle={Workshop on Medical Image Learning with Limited and Noisy Data},
  pages={246--256},
  year={2023},
  organization={Springer}
}

@article{yousefirizi2022novel,
	author = {Yousefirizi, Fereshteh and Jha, Abhinav and Ahamed, Shadab and Bloise, Ingrid and O, Joo Hyun and Sehn, Laurie and Savage, Kerry and Uribe, Carlos and Rahmim, Arman},
	title = {A novel loss function for improved deep learning-based segmentation: Implications for TMTV computation},
	volume = {63},
	number = {supplement 2},
	pages = {2588--2588},
	year = {2022},
	publisher = {Society of Nuclear Medicine},
	issn = {0161-5505},
	URL = {https://jnm.snmjournals.org/content/63/supplement_2/2588},
	eprint = {https://jnm.snmjournals.org/content},
	journal = {Journal of Nuclear Medicine}
}

@article{ahamed2026comprehensive,
  title={Comprehensive framework for evaluation of deep neural networks in detection and quantification of lymphoma from PET/CT images: clinical insights, pitfalls, and observer agreement analyses},
  author={Ahamed, Shadab and Xu, Yixi and Kurkowska, Sara and Gowdy, Claire and HO, Joo and Bloise, Ingrid and Wilson, Don and Martineau, Patrick and B{\'e}nard, Fran{\c{c}}ois and Yousefirizi, Fereshteh and others},
  journal={Physica Medica},
  volume={148},
  pages={105832},
  year={2026},
  publisher={Elsevier}
}

@article{alasmawi2026advanced,
  title={Advanced Tumor Segmentation in PET/CT Imaging: A Training Strategy Study with nnU-Net for AutoPET III},
  author={Alasmawi, Hussain},
  journal={arXiv preprint arXiv:2605.08161},
  year={2026}
}

@article{murugesan2023improving,
  title={Improving lesion segmentation in FDG-18 whole-body PET/CT scans using multilabel approach: Autopet II challenge},
  author={Murugesan, Gowtham Krishnan and McCrumb, Diana and Brunner, Eric and Kumar, Jithendra and Soni, Rahul and Grigorash, Vasily and Moore, Stephen and Van Oss, Jeff},
  journal={arXiv preprint arXiv:2311.01574},
  year={2023}
}

@article{wang2024dual,
  title={Dual channel cw nnu-net for 3d pet-ct lesion segmentation in 2024 autopet iii challenge},
  author={Wang, Ching-Wei and Su, Ting-Sheng and Liu, Keng-Wei},
  journal={arXiv preprint arXiv:2409.07144},
  year={2024}
}

@article{kalisch2024autopet,
  title={Autopet III challenge: Incorporating anatomical knowledge into nnUNet for lesion segmentation in PET/CT},
  author={Kalisch, Hamza and H{\"o}rst, Fabian and Herrmann, Ken and Kleesiek, Jens and Seibold, Constantin},
  journal={arXiv preprint arXiv:2409.12155},
  year={2024}
}

@article{chung2014fdg,
  title={FDG PET/CT metabolic tumor volume and total lesion glycolysis predict prognosis in patients with advanced lung adenocarcinoma},
  author={Chung, Hyun Woo and Lee, Kye Young and Kim, Hee Joung and Kim, Wan Seop and So, Young},
  journal={Journal of cancer research and clinical oncology},
  volume={140},
  number={1},
  pages={89--98},
  year={2014},
  publisher={Springer}
}

@article{gatidis2022whole,
  title={A whole-body FDG-PET/CT dataset with manually annotated tumor lesions},
  author={Gatidis, Sergios and Hepp, Tobias and Fr{\"u}h, Marcel and La Foug{\`e}re, Christian and Nikolaou, Konstantin and Pfannenberg, Christina and Sch{\"o}lkopf, Bernhard and K{\"u}stner, Thomas and Cyran, Clemens and Rubin, Daniel},
  journal={Scientific Data},
  volume={9},
  number={1},
  pages={601},
  year={2022},
  publisher={Nature Publishing Group UK London}
}

@article{rokuss2024fdg,
  title={From FDG to PSMA: A Hitchhiker's Guide to Multitracer, Multicenter Lesion Segmentation in PET/CT Imaging},
  author={Rokuss, Maximilian and Kovacs, Balint and Kirchhoff, Yannick and Xiao, Shuhan and Ulrich, Constantin and Maier-Hein, Klaus H and Isensee, Fabian},
  journal={arXiv preprint arXiv:2409.09478},
  year={2024}
}

@article{dexl2026autopet3,
  title={The autoPET3 Challenge: Automated Lesion Segmentation in Whole-Body PET/CT Multitracer Multicenter Generalization},
  author={Dexl, Jakob and Jeblick, Katharina and Mittermeier, Andreas and Schachtner, Balthasar and St{\"u}ber, Anna Theresa and Topalis, Johanna and Rokuss, Maximilian and Isensee, Fabian and Maier-Hein, Klaus H and Kalisch, Hamza and others},
  journal={arXiv preprint arXiv:2605.05775},
  year={2026}
}

@article{mcdonald2017assessment,
  title={Assessment of total lesion glycolysis by 18F FDG PET/CT significantly improves prognostic value of GEP and ISS in myeloma},
  author={McDonald, James E and Kessler, Marcus M and Gardner, Michael W and Buros, Amy F and Ntambi, James A and Waheed, Sarah and van Rhee, Frits and Zangari, Maurizio and Heuck, Christoph J and Petty, Nathan and others},
  journal={Clinical Cancer Research},
  volume={23},
  number={8},
  pages={1981--1987},
  year={2017},
  publisher={American Association for Cancer Research}
}

@article{maier2024metrics,
  title={Metrics reloaded: recommendations for image analysis validation},
  author={Maier-Hein, Lena and Reinke, Annika and Godau, Patrick and Tizabi, Minu D and Buettner, Florian and Christodoulou, Evangelia and Glocker, Ben and Isensee, Fabian and Kleesiek, Jens and Kozubek, Michal and others},
  journal={Nature methods},
  volume={21},
  number={2},
  pages={195--212},
  year={2024},
  publisher={Nature Publishing Group US New York}
}

@article{reinke2024understanding,
  title={Understanding metric-related pitfalls in image analysis validation},
  author={Reinke, Annika and Tizabi, Minu D and Baumgartner, Michael and Eisenmann, Matthias and Heckmann-N{\"o}tzel, Doreen and Kavur, A Emre and R{\"a}dsch, Tim and Sudre, Carole H and Acion, Laura and Antonelli, Michela and others},
  journal={Nature methods},
  volume={21},
  number={2},
  pages={182--194},
  year={2024},
  publisher={Nature Publishing Group US New York}
}

@article{gatidis2024results,
  title={Results from the autoPET challenge on fully automated lesion segmentation in oncologic PET/CT imaging},
  author={Gatidis, Sergios and Fr{\"u}h, Marcel and Fabritius, Matthias P and Gu, Sijing and Nikolaou, Konstantin and Foug{\`e}re, Christian La and Ye, Jin and He, Junjun and Peng, Yige and Bi, Lei and others},
  journal={Nature Machine Intelligence},
  volume={6},
  number={11},
  pages={1396--1405},
  year={2024},
  publisher={Nature Publishing Group UK London}
}
}

\end{document}